\RequirePackage{fix-cm}
\documentclass[twocolumn]{svjour3}          
\smartqed  

\usepackage{graphicx}
\usepackage{amsmath,amsfonts,amssymb}
\usepackage{graphicx}
\usepackage{wrapfig}
\usepackage{url}
\usepackage{ctable}
\usepackage{multirow}
\usepackage[titletoc]{appendix}
\usepackage{lipsum}
\usepackage{mdframed}
\usepackage{fancyvrb}
\usepackage{tabu}
\usepackage{tabularx}
\usepackage{booktabs}
\usepackage{multirow}
\usepackage{xspace}
\usepackage{enumitem}
\usepackage{algorithm}
\usepackage{algorithmic}

\usepackage{dsfont}

\newcommand{\bfU}{\textbf{u}}

\newcommand{\bftheta}{{\boldsymbol{\theta}}}

\newcommand{\R}{\mathds{R}}

\journalname{}

\begin{document}

\title{Multilevel Initialization for Layer-Parallel Deep Neural Network Training 
}


\author{Eric C. Cyr \and Stefanie G\"{u}nther \and Jacob B. Schroder}



\institute{Eric C. Cyr \at
           Sandia National Laboratories\\
           \email{eccyr@sandia.gov}
           \and
           Stefanie G\"unther \at
           Lawrence Livermore National Laboratory\\
           \email{guenther5@llnl.gov}
           \and
           Jacob B. Schroder \at
           Dept. of Mathematics and Statistics, University of New Mexico\\
           \email{jbschroder@unm.edu}
}

\date{Received: date / Accepted: date}

\maketitle

\begin{abstract}
This paper investigates multilevel initialization strategies for training very deep neural networks with a layer-parallel multigrid solver. The scheme is based on the continuous interpretation of the training problem as a problem of optimal control, in which neural networks are represented as discretizations of time-dependent ordinary differential equations. A key goal is to develop a method able to intelligently initialize the network parameters for the very deep networks enabled by scalable layer-parallel training.  To do this, we apply a refinement strategy across the time domain, that is equivalent to refining in the layer dimension.  The resulting refinements create deep networks, with good initializations for the network parameters coming from the coarser trained networks.  We investigate the effectiveness of such multilevel ``nested iteration" strategies for network training, showing supporting numerical evidence of reduced run time for equivalent accuracy.  In addition, we study whether the initialization strategies provide a regularizing effect on the overall training process and reduce sensitivity to hyperparameters and randomness in initial network parameters.

\end{abstract}
\section{Introduction}
\label{intro}


We consider learning problems that are based on the continuous neural network formulation as in \cite{Haber_2017,ChRuBeDu2018}, and note that there has been considerable interest in related ODE- and PDE-inspired formulations as of late \cite{weinan2017proposal,chang2018reversible,chaudhari2018deep,lu2017beyond,ruthotto2018deep}. 
In this setting, the neural network is cast as a time-dependent ordinary differential equation (ODE) which describes the flow of data elements $y_k \in \R^{n_f}, k=1,\dots,s$ (the \textit{feature} vectors) through the neural network as
\begin{align}
  \partial_t u_k(t) &= F(u_k(t), \theta(t)) \quad \forall t \in (0,T) \label{network_flow}\\
  u_k(0) &= L_{in} y_k \label{network_init}.
\end{align}
Here, $u_k(t)\in R^w$ describes the state of a neural network of width $w$, and $\theta(t)\in\R^p$ represents the network parameters (weights). $L_{in}$ maps the input feature vectors $y_k$ to the network dimension; the right-hand-side $F$ then determines the flow of the data element through the network. $F$ typically consists of an affine transformation that is parameterized by $\theta(t)$, and a nonlinear transformation that is applied element-wise using an {activation} function, i.e. 
\begin{align}
  F(u_k(t),\theta(t)) = \sigma(W(\theta(t)) u_k(t) + b(\theta(t))),
\end{align}
where $W(\theta(t))$ is a linear transformation, such as a convolution, applied to the state $u_k(t)$, $b(\theta(t))$ is a bias added to the state, and $\sigma\colon \R \to \R$ denotes the nonlinear activation function applied componentwise, such as the ReLU activation $\sigma(x) = \max(x,0)$.

In contrast to traditional neural networks that transform the network state at prescribed layers, the continuous formulation prescribes the rate of change of the network state using an ODE parameterized by $\theta(t)$. This control function $\theta(t)$ is to be learned during training by solving a constrained optimization problem. For example, in supervised learning, the optimization problem aims to match the network output $u_k(T)$ to a desired output $c_k \in \R^{n_c}$ that is given from the training data set:
\begin{align} \label{eq:training}
  \min_{\theta} \, &\frac 1 s \sum_{k=1}^s \ell(  u_k(T), c_k) + \int_0^T R(\theta(t)) \, \mathrm{d} t \\
  \text{s.t. equations } &\eqref{network_flow}-\eqref{network_init} \mbox{ are satisfied } \forall \, k=1,\dots, s.
\end{align}
Here, $\ell$ denotes a loss function that measures the distance of the network output at final time $T$ to the desired output, and $R$ denotes a regularization term, such as a Tikhonov regularization on $\theta(t)$ and/or its time derivative, that stabilizes the optimization problem numerically.
Training is generally considered successful, if the network parameters $\theta(t)$ generalize well to new, previously unseen data,
which is represented as a validation data set.

To solve the neural network flow numerically, classical time-integration schemes are applied that discretize the ODE in equation \eqref{network_flow} on a given time-grid $0=t_1<\dots< t_N=T$, and solve the resulting equations one after another for each time step $t_i$. In this scenario, each time-step is associated with one layer of a classical artificial neural network, leading to the traditional network propagation with $u^n_k \approx u_k(t^n)$ and $\theta^n\approx \theta(t^n)$ being the network state and weights at layer $n$, respectively. In fact, it has been observed that many state-of-the-art artificial neural networks can be interpreted as discrete time-integration schemes of the parametrized ODE in equation \eqref{network_flow} (see e.g. \cite{LuZhLiDo2017}). For example, an explicit Euler scheme to discretize \eqref{network_flow} gives
\begin{align}
  u^{n+1}_k = u^n_k + hF(u_k^n, \theta^n), \label{eq:resnet}
\end{align}
which resembles the classical ResNet~\cite{HeZaReSu2016} with an additional time-step size $h>0$.  We note that $h=1$ in the classical ResNet.

The continuous network formulation opens the door for
leveraging various optimization schemes and optimal control theory that has been developed for ODE and PDE constraints \cite{Li1971,Tr2010,BiGhHeBl2003}.
In particular, stability analysis of explicit time-integration schemes suggests favoring many-layer networks utilizing a small step size $h$ in order to ensure a numerically stable network propagation. Networks based on numerical time-integration schemes can therefore easily involve hundreds or thousands of discrete layers (time-steps) \cite{GuRuScCyGa2018,ChRuBeDu2018,Haber_2017}. 

However, training such huge networks comes with both numerical and computational challenges. First of all, the serial nature of time evolution creates a barrier for parallel scalability. If network states are propagated through the network in a serial manner, as is done with classical training methods, an increase in the number of layers (i.e. more time steps, and smaller steps sizes) results in an equally larger time-to-solution. In order to address this serial bottleneck, a layer-parallel multigrid scheme has been developed in \cite{GuRuScCyGa2018} to replace the serial forward and backward network propagation. In this scheme, the network layers are distributed onto multiple compute units, and an iterative multigrid method is applied to solve the network propagation equations inexactly in parallel, across the layers.  This iterative scheme converges to the same solution as serial network propagation.  The result is that runtimes remain nearly constant when the numbers of layers and compute resources are increased commensurately (weak scaling), and that runtimes can be significantly reduced for a fixed number of layers and increasing compute resources (strong scaling). The layer-parallel multigrid method will be summarized in Section \ref{sec:MGRIT}.

In addition, many-layer networks increase the complexity of the underlying optimization problem. In particular, considering more and more layers and hence more and more network parameters creates highly non-convex optimization landscapes which require proper initialization in order to be trained effectively. A number of schemes have been developed for initializing plain and residual neural networks. Commonly implemented techniques include Glorot~\cite{glorot10} and He~\cite{he2015delving}; however, this is still an active area of research, with new methods being proposed (e.g.~\cite{hanin2018start,humbird2018deep,boxinit2019}). 

In this paper, we investigate a multilevel network initialization strategy that successively increases the number of network layers during layer-parallel multigrid training. The first use of such a multilevel (or ``cascadic") initialization strategy for an ODE network was done in the context of layer-serial training in \cite{Haber_2017}.  Here, a sequence of increasingly deeper network training problems are solved, and each new deeper network is initialized with the interpolated learned parameters from the previous coarser network. We refer to this process as a ``nested iteration", because of this terminology's long history. Nested iteration broadly
describes a process (originally for numerical PDEs/ODEs) that starts by finding the solution for a relatively coarse problem representation, where the solution
cost is relatively cheap.  This coarse solution is then used as an inexpensive initial guess for the same problem, 
but at a finer resolution. Nested iteration was first discussed in a multigrid context at least in 1981
\cite{hackbusch1981convergence}, 
and the concept of nested iterations for solving numerical PDEs goes back at least to 
1972 \cite{kronsjo1972design,Kr1975}.  Nested iteration is often called ``full multigrid" in the
multigrid context, and famously provides optimal $O(n)$ solutions to some elliptic problems with $n$ unknowns 
\cite{BrHeMc2000,TrOo2001}. Nested iteration, especially when only regions with large error are marked for refinement, 
can lead to remarkably efficient solvers \cite{AdMaMcNoRuTa2011,DeMa2008}.  We will take advantage of this
inherent efficiency to cheaply find good initializations for ODE networks, by successively refining
the time grid of the ODE network, adding more and more time-grid levels to the multigrid hierarchy.


In this work, we put the nested iteration (or cascadic) initialization idea into the context of an iterative layer-parallel multigrid solver, which re-uses the coarser grid levels during multigrid cycling. We investigate two interpolation strategies (constant and linear) and their influence on the layer-parallel multigrid convergence and training performance. We further investigate the effect of multilevel initialization as a regularization force, with the desire that deep network training can become less sensitive towards hyperparameters and variation in network parameter initializations.  In other words, the goal is that with a better initial guess for the network parameters, the training process becomes more robust and less sensitive.

\section{Methods: Layer-Parallel Training and Nested Iteration}
\label{sec:methods}

\subsection{Layer-Parallel Training via Multigrid}
\label{sec:MGRIT}

In this section, we summarize the layer-parallel training approach as presented in \cite{GuRuScCyGa2018}. For a full description of the multigrid training scheme, the reader is referred to the original paper, and references therein. 

At the core of the layer-parallelization technique is a parallel multigrid algorithm that is applied to the discretized ODE network, replacing the serial forward (and backward) network propagation. 
Consider the discretized ODE network propagation to be written as 
\begin{align}
  u^{n+1} = \Phi_h(u^n, \theta^n) \label{eq:forward_prop}.
\end{align}
For example, one can choose $\Phi(u^n, \theta^n) = u^n + hF(u^n, \theta^n)$, thus letting $\Phi$ denote the right hand side of \eqref{eq:resnet} for a ResNet with step size $h$ and $h := t^{n+1} - t^n$ on a uniform time grid.
Classical sequential training solves \eqref{eq:forward_prop} forward in time, starting from $u^0 = L_{in}y$ for either a general feature vector $y$, or for a batch of feature vectors $y = \{y_k\}_{k\in S\subset\{1,\dots,s\}}$, finally stopping at the network output $u^N$.
On the other hand, the layer-parallel multigrid scheme solves \eqref{eq:forward_prop} by simultaneously computing all layers with an iterative nonlinear multigrid scheme (FAS, \cite{BrHeMc2000}) applied to the network layer domain. This multigrid scheme essentially computes inexact forward- and backward-propagations in an iterative fashion, such that the process converges to the layer-serial solution.

To compute these inexact propagations, a hierarchy of ever coarser layer-grid levels is created, which in turn accelerate convergence to the layer-serial solution on the finest level.  A coarser level is created by assigning every $c$-th layer to the next coarser level, giving the step size $h_l := h c^l$ at the $l$-th level, with level $l=0$ being the finest. For each level, this assignment results in a partitioning of the layers into $F$- (fine-grid) and $C$- (coarse-grid) layers. 

\begin{figure}\label{fig:FCrelaxation}
  \center
  \includegraphics[width=.9\columnwidth]{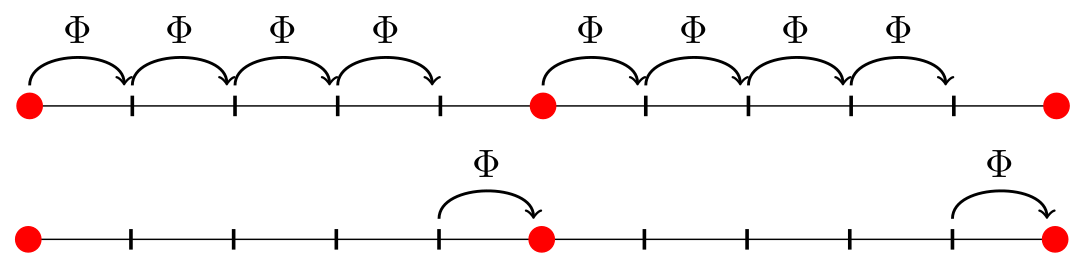}
  \caption{$F$-layer (top) and $C$-layer (bottom) smoothing operations for a coarsening factor of $c=5$. Graphic taken from \cite{GuRuScCyGa2018}.}
\end{figure}

Each multigrid iteration then cycles through the different layer-grid levels while applying smoothing operations to update (improve) the network state at $F$- and $C$- layers, which can be updated in parallel in an alternating fashion. Each $F$- or $C$-layer smoothing operation consists of an application of the layer propagator $\Phi_{h_l}$, using the step size $h_l$ of the current grid, to update the $F$- or $C$- layer states, see Figure \ref{fig:FCrelaxation}.  Additionally, each coarse level has a correction term that involves the residual of states between successive layer-grid levels, which is a part of the FAS method. 
Typically, $FCF$-smoothing is applied on each grid level which refers to a sucessive application of $F$-smoothing, then $C$-smoothing, then $F$-smoothing again. Note that these smoothing operations are highly parallel as $F$- and $C$-point sweeps are applied locally on each layer interval and independently from each other. Serial propagation only occurs on the coarsest level, where the problem size is trivial. 

At convergence, the layer-parallel multigrid solver reproduces the same network output as serial propagation (up to a tolerance). However it enables concurrency across the layers. This concurency creates a cross-over point in terms of computational resources, after which the layer-parallel multigrid approach provides speedup over the layer-serial computation. 

The same layer-parallel multigrid strategy can be applied to parallelize both the forward propagation and the backpropagation to compute the gradient with respect to the network weights $\theta^n$. For the latter, the layer propagator at each grid level propagates partial derivatives of $\Phi_{h_l}$ with respect to $u^n$ and $\theta^n$ backwards through the time domain (through the layers), again locally and in parallel.

The layer-parallel solvers could in principle be substituted for forward- and backpropagation in any gradient-based optimization scheme for solving \eqref{eq:training} when training.  One would only replace the sequential forward- and backpropagation with the layer-parallel multigrid iterations. However due to its iterative nature, the layer-parallel scheme is very well suited for simultaneous optimization approaches that utilize inexact gradient information to update the network weights. In \cite{GuRuScCyGa2018}, the simultaneous One-shot method is applied, which performs a fixed number of $d$ multigrid iterations for the network state and its derivative, with $d$ being as low as $2$, before each update of the network weights. Therein, a deterministic optimization approach has been applied, involving second order Hessian information. However the use of this approach with stochastic optimization, e.g., stochastic gradient descent method (SGD), is possible, although not yet tested numerically. 



\subsection{Nested Iteration (Multilevel) Initialization of Deep Neural Networks}
\label{sec:nested_iter}

\begin{figure}\label{fig:NI}
\centering{
\includegraphics[width=0.49\textwidth]{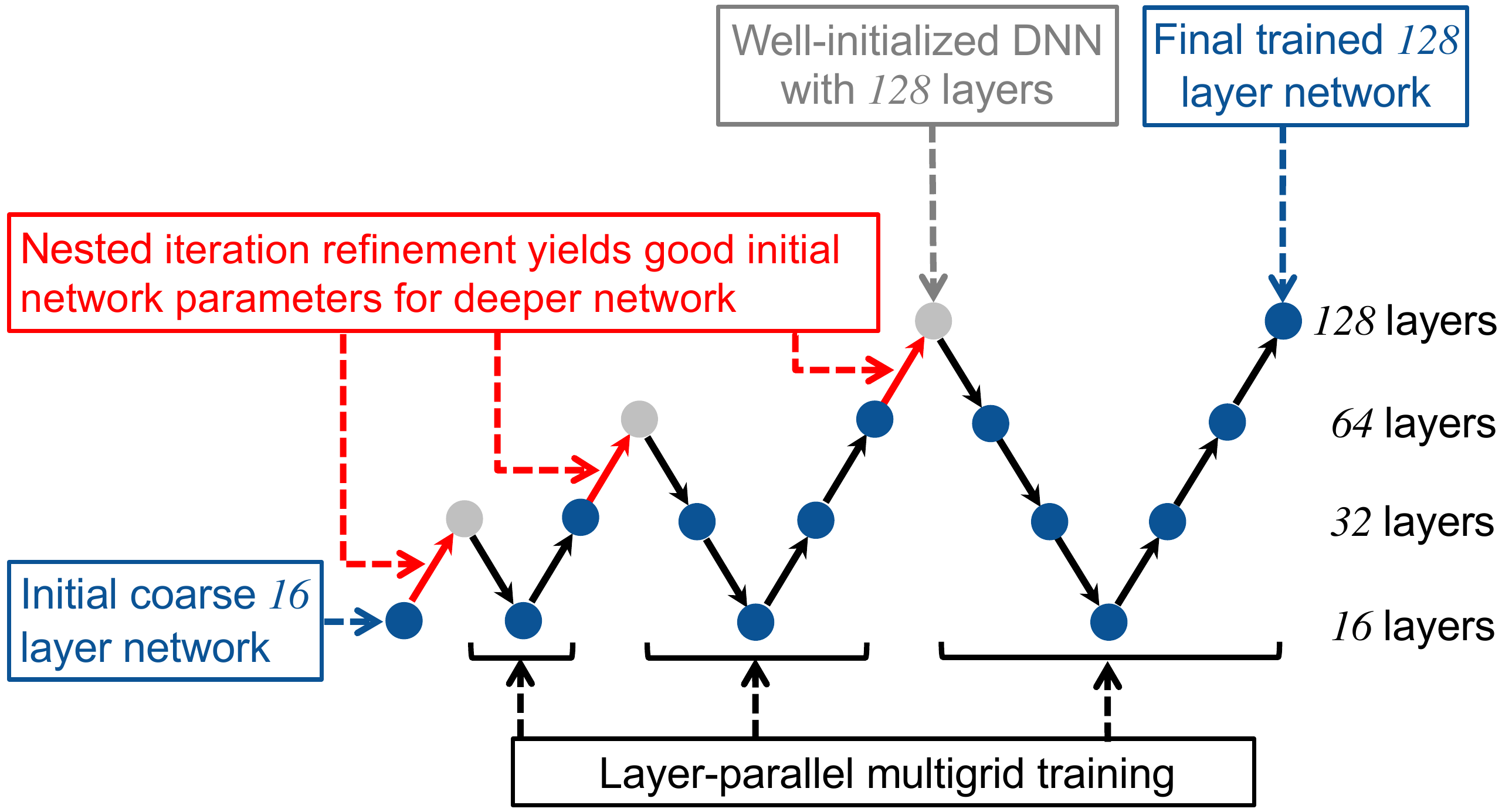}}
\caption{Nested iteration algorithm, starting with a coarse 16 layer network, and then carrying out 3 refinements 
to reach a 128 layer network.  Refinements are in red, and the black arrows depict the layer-parallel multigrid 
training, which is not to be confused with nested iteration cycling.}
\end{figure}

Our proposed answer for initializing deep networks is nested iteration, where a trained
coarse network, with fewer time-steps, is interpolated to initialize a finer network, with more time-steps.  
Our nested iteration algorithm is depicted in Figure \ref{fig:NI} and Algorithm \ref{alg:NI}, with the following notation.
Let the total number of nested iteration levels be $L$, where $L=0$ is the finest level (i.e., largest network)
and the superscript $^{(l)}$ denotes quantities on nested iteration level $l$, e.g., 
$\bfU^{(L-1)}$ and $\bftheta^{(L-1)}$ are the coarsest-level state and control (weights and biases) variables \emph{for all time steps (layers)}.  
Additionally, let $\{ m^{(l)} \}$ be a set of size $L$ containing level dependent optimization iteration counts, 
e.g., $m^{(0)}$ denotes the number of optimization iterations to carry out on the final, finest level $0$.  
Line \ref{alg:lpt} of Algorithm \ref{alg:NI} represents one optimization iteration of the layer-parallel training approach to update the current weights $\bftheta^{(l)}$, applying $d$ inner layer-parallel multigrid iterations.

Finally, we define the interpolation operator $P$ to interpolate the weights and biases $\bftheta^{(l)}$ to the next
finer level $(l-1)$.  The interpolation here is a uniform refinement in time with refinement factor 2.
That is, there are exactly twice as many time-steps on the finer grid $(l-1)$ as on the coarser grid $(l)$.
Thus, if $L=4$ and the initial number of layers were $16$, the final network would have $128$ layers, as depicted in Figure \ref{fig:NI}.
We allow for two types of interpolation, piece-wise constant in time and linear in time, and thus define $P$ as
\begin{align}
\quad \bftheta^{(l-1)}      &= P^{(l)} \bftheta^{(l)}, \; \mbox{where} \\
    \bftheta^{(l-1)}_{2n}   &= \bftheta^{(l)}_n,\; 
                           \mbox{and} \nonumber \\
    \bftheta^{(l-1)}_{2n+1} &= \bftheta^{(l)}_n, \mbox{if piece-wise constant} \nonumber \\
    \bftheta^{(l-1)}_{2n+1} &= \frac 1 2 (\bftheta^{(l)}_n + \bftheta^{(l)}_{n+1}), \mbox{if linear,} \nonumber
\end{align}
for $n=0,\dots, N^{(l)}$ with $N^{(l)}$ being the number of layers on nested iteration level $l$.

\begin{algorithm}
  \caption{nested\_iter($\bfU^{(L-1)}$, $\bftheta^{(L-1)}$, $L$, $\{m^{(l)}\}$ ) } 
  \label{alg:NI}
  \begin{algorithmic}[1]
     \STATE $\triangleright$ \textit{Loop over nested iter. levels, then optimization iter.}
     \STATE Initialize $\bfU^{(L-1)}$, $\bftheta^{(L-1)}$ 
     \FOR{$l = L-1,\; l > 0,\; l\; \mbox{-=} 1$}
        \FOR{$i = 0,\; i < m^{(l)},\; i\; \mbox{+=} 1$}
            \STATE $\bfU^{(l)}, \bftheta^{(l)} \leftarrow LPT(\bfU^{(l)}, \bftheta^{(l)}, d)$  \hfill $\triangleright$ \textit{LPT: Layer- $\quad$}\label{alg:lpt}
            \STATE \hfill \textit{parallel training}
        \ENDFOR
        \STATE $\bftheta^{(l-1)} = P^{(l)} \bftheta^{(l)}$ \hfill $\triangleright$ \textit{Interpolate} \label{line:interp}
     \ENDFOR
     \RETURN $\bftheta^{(0)}$  \hfill $\triangleright$ \textit{Return finest level weights}
  \end{algorithmic}
\end{algorithm}

The weights $\bftheta^{(L-1)}$ on the coarsest nested iteration level are initialized according to the overall
hyperparameters, e.g., zero weights for internal layers, and random weights from a certain distribution for the
opening and closing layers.  These strategies are discussed in the results Section \ref{results}.  Additionally,
the choice between linear and piece-wise constant interpolation now becomes another hyperparameter to choose.
In our experiments, both strategies performed similarly, in terms of multigrid convergence and training and validation
accuracy.  Thus, we report results only for piece-wise constant interpolation in Section \ref{results}.

One key detail in Algorithm \ref{alg:NI} is the choice of the number of inner layer-parallel multigrid iterations $d$. It has been observed numerically, that a small choice of $d$ during early optimization iterations can lead to steep drops 
in the training and validation accuracy 
after the interpolation step in line \ref{line:interp}.  
Rather than being related to the interpolation itself, we observed that 
the true cause of these steep drops
were inaccurate multigrid solves which lead to big errors in the gradient, and thus poor updates 
to the weights.  We therefore enforce more multigrid iterations immediately after interpolation, in particular we choose $d=10$ for the first $3$ optimization iterations after interpolation to the new grid level, and $d=2$ after that (i.e. after $i\geq3$).
In general, we 
recommend to apply enough multigrid iterations that ensure to drop the multigrid error below a relative error tolerance, such as guaranteeing a relative error drop of the multigrid residual of 
4 orders of magnitude.  It is important to note that this issue does not occur when using nested iteration 
with sequential training, as in \cite{Haber_2017}.

Other possible enhancements to the algorithm, such as refinement by other factors than 2 and other 
interpolation formulas are topics for future research.

%

\section{Results}
\label{results}

We use two machine learning problems to demonstrate the performance of the nested iteration algorithm: 

\begin{enumerate}
\item ``Peaks'' example: \\
\noindent The first problem is referred to as ``peaks'', and suggested in~\cite{Haber_2017}. The task is to classify particles as members of five distinct sets. We train on $s=5000$ data points consisting of particle positions ${y}_k \in [-3,3]^2$, while membership in the sets is defined by a vector of probabilities $c_k \in \mathbb{R}^5$ (unit vectors). 
\item Indian Pines: \\
\noindent The second example is a hyperspectral image segmentation problem using the Indian Pines image data set~\cite{indianpinesdata}. The classification task for this problem is to assign each pixel of a $145\times 145$ pixel image to one of $16$ classes representing the type of land-cover (e.g. corn, soy, etc...), based on $220$ spectral reflectance bands representing portions of the electromagnetic spectrum. We choose $s=1000$ pixels for training. 
\end{enumerate}

For both examples, we choose a ResNet architecture as in \eqref{eq:resnet}, with a ReLU activation function $\sigma$ that is smoothed around zero. The linear transformations inside each layer consist of a dense matrix of network weights.

In order to provide a fair basis for comparison between the nested iteration and non-nested iteration training simulations we introduce a common ``work unit.'' In all examples below, the work unit is defined as the average wall-clock run time for each iteration of the non-nested iteration on the fine grid. Thus the number of work units required for a non-nested iteration is equal to the number of optimization iterations. For the nested iteration, this rescaling of the run time provides a common basis for comparison to the non-nested iteration. 
In addition, the metric used for comparison below is the \emph{validation accuracy}. This is measured by withholding a subset of the data from training and checking the performance of inference on those values. The percentage (ranging from $0\%$ to $100\%$), of correctly classifying members of that data set is the validation accuracy.

\subsection{Peaks Example}

Two versions of the peaks problem are run with residual networks of width $5$ and $8$. For both the nested and non-nested cases, $16$ processors are used for the inner layer-parallel solve. 
The nested iteration is run with a schedule starting with $m^{(2)}=200$ coarse steps, $m^{(1)}=125$ steps, and $m^{(0)}=75$ fine level steps.
For non-nested iteration $188$ optimization steps are taken (the number of steps was chosen so the run time of non-nested was nearly the same as nested iteration).  The final simulation time for the peaks problem is set to $T=5$.

A challenge facing neural networks usage is the range of parameters associated with both their design and training. In addition, due to the variability of the loss surface, the initial guess for the controls and state variables can have a dramatic impact on the quality of the training. 
The tables in Figure~\ref{fig:peaks-tables} show statistics for $12$ independent training runs, for each of $4$ sets of parameters, yielding a total of $48$ uniquely trained neural networks.
The parameters selected for this study were the initial magnitude of the randomly initialized weights denoted $w_i$, and the Tikhonov regularization parameter denoted $\gamma_T$. Based on a larger hyperparameter scan, we selected two values for each parameter that yield the greatest validation accuracy for nested iteration and non-nested iteration. As a result $w_i\in\{0.0,10^{-6}\}$ and $\gamma_T\in\{10^{-5},10^{-7}\}$, yielding a total of four parameters.

The top table shows results for a network with width $5$, while the bottom shows the results for width $8$. In both cases, there are $64$ residual layers, plus an opening and classification layer.
The tables show that the nested iteration achieves better validation accuracy on average for both network configurations. In addition, there is less variation as measured by both the standard deviation and the range of extrema using nested iteration. We attribute this to the use of a sequence of coarse grid problems to define an improved initial guess for the fine grid. At the coarse level, because of the reduced parameter space, the variation seen in the objective surface is potentially not as large. When only the fine simulation is used in non-nested iteration, the likelihood that the training algorithm gets stuck in a local minima early in the process is likely increased. In effect, the nested iteration is behaving like a structural regularization approach for the objective surface. 

\begin{figure}
\begin{center}
\begin{tabular}{ l|cc }
  & \multicolumn{2}{|c}{$5$ Channel} \\
 \hline
 & Nested & Non-Nested \\
 \hline
 Mean & 86.7\% & 85.0\%  \\ 
 Median & 88.0\% & 88.5\%  \\ 
 Max & 97.0\% & 95.0\%  \\ 
 Min & 66.0\% & 20.0\% \\
 Std. Dev & 7.69\% & 11.7\%
\end{tabular}
\end{center}

\begin{center}
\begin{tabular}{ l|cc }
  & \multicolumn{2}{|c}{$8$ Channel} \\ 
 \hline
 & Nested & Non-Nested \\
 \hline
 Mean  & 92.3\% & 90.7\% \\ 
 Median  & 94.0\% & 91.8\% \\ 
 Max  & 99.0 \% & 96.5\% \\ 
 Min  & 72.5 \% & 57.0\% \\
 Std. Dev & 5.18 \% & 6.08 \% 
\end{tabular}
\end{center}
\caption{These tables show the statistical variation of the peaks example run using a scan over $4$ sets of hyperparameters, and $12$ training runs. For both the $5$ channel and $8$ channel peaks problem the nested iteration demonstrates less sensitivity to hyperparameter choice and initialization then the non-nested iteration. \label{fig:peaks-tables}}
\end{figure}

\begin{figure} 
\centering{
\includegraphics[width=0.4\textwidth]{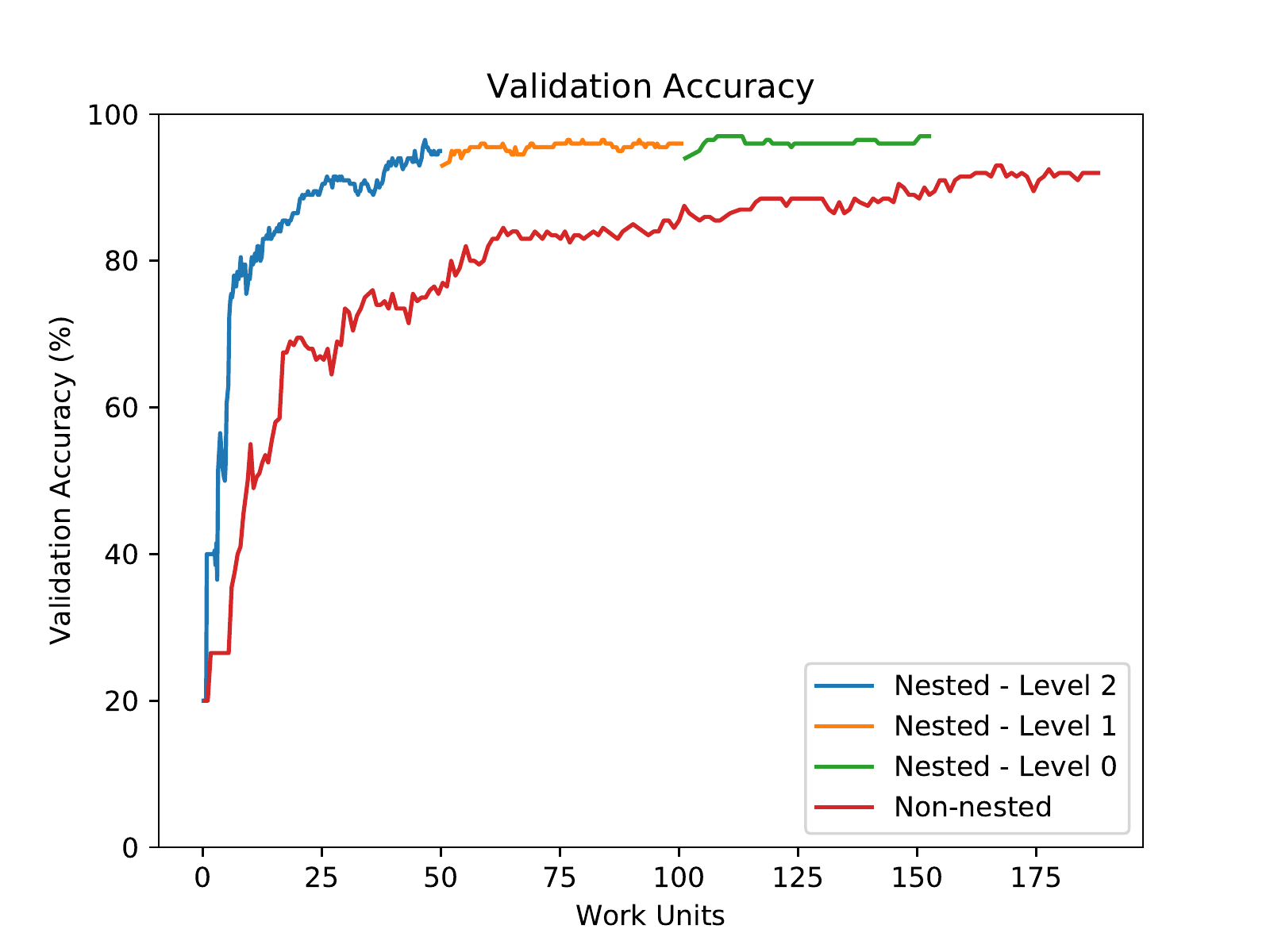}}
\centering{
\includegraphics[width=0.4\textwidth]{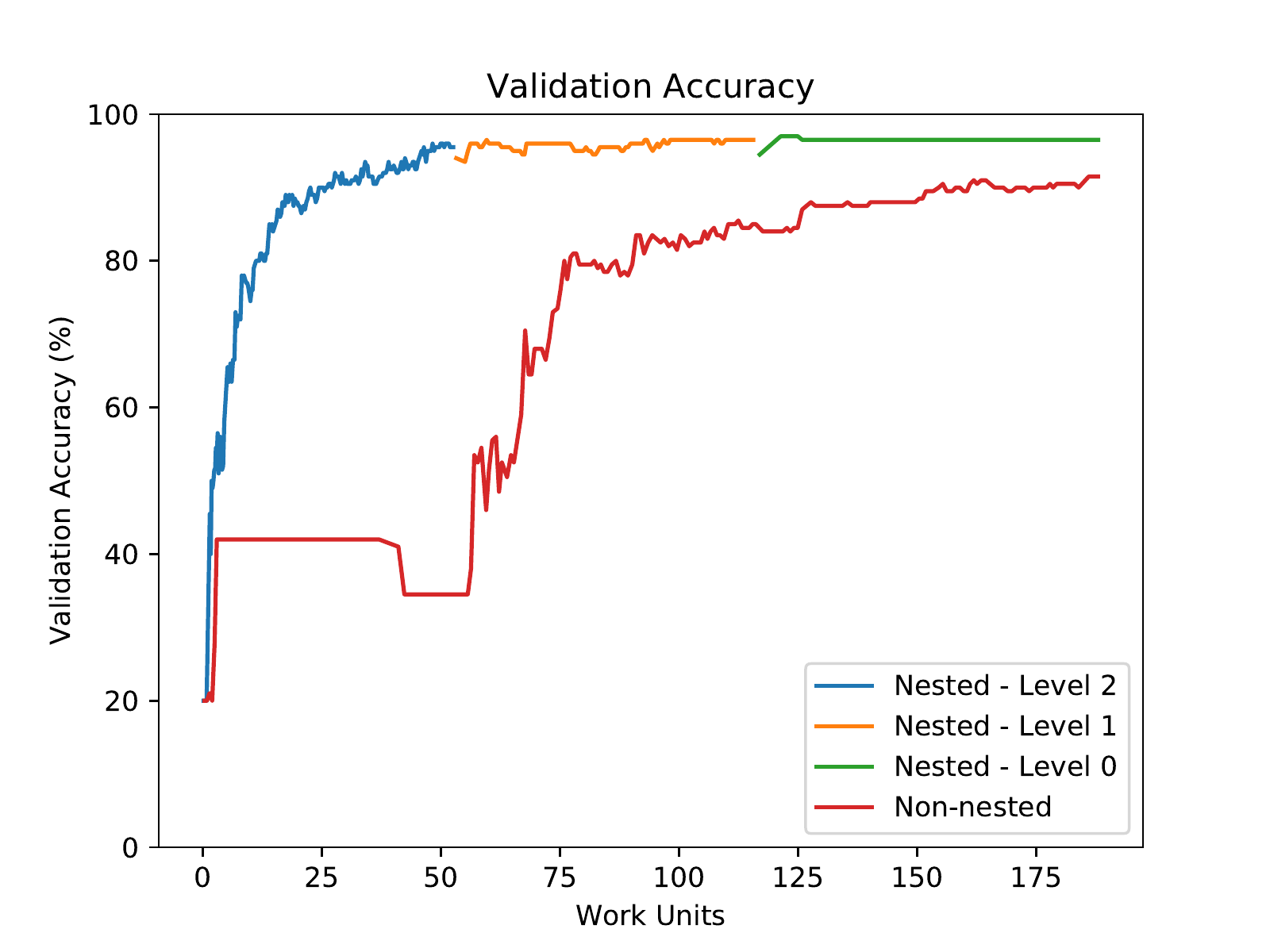}}
\caption{Validation accuracy as a function of computational work units for the peaks problem. The top image shows a network containing $5$ channels, while the network in the bottom image contains $8$. The validation accuracy is uniformly larger for the nested iteration.\label{fig:peaks-train-images}}
\end{figure}

Figure~\ref{fig:peaks-train-images} shows the validation accuracy of the peaks problem as a function of work units for the individual best runs of nested iteration and non-nested iteration. The non-nested iteration corresponds to a single red line. The three levels of the nested iteration are plotted in a sequence of colors that show the achievement for each level of the algorithm. Again this is a function of work units, so these levels are scaled relative to the cost of a single fine level optimization step (some variability may occur due to the use of a backtracking algorithm). 

In both the top image (for width $5$ residual networks) and the lower image (width $8$), the nested iteration has clearly superior validation accuracy as compared to the non-nested iteration. Moreover, the accuracy achieved for \emph{any} number of work units is larger with nested iteration. 

\subsection{Indian Pines Example}
For the Indian Pines example, we use a residual neural network with width $220$, that contains at the fine level $128$ residual layers, plus opening and classification layers. For nested iteration, we use three levels with the coarsest having $32$ residual layers. The final simulation time is $T=5$. All runs were performed on $32$ processors, implying the coarse grid contains $1$ residual layer per processor. 

\begin{figure}
\centering{
\includegraphics[width=0.4\textwidth]{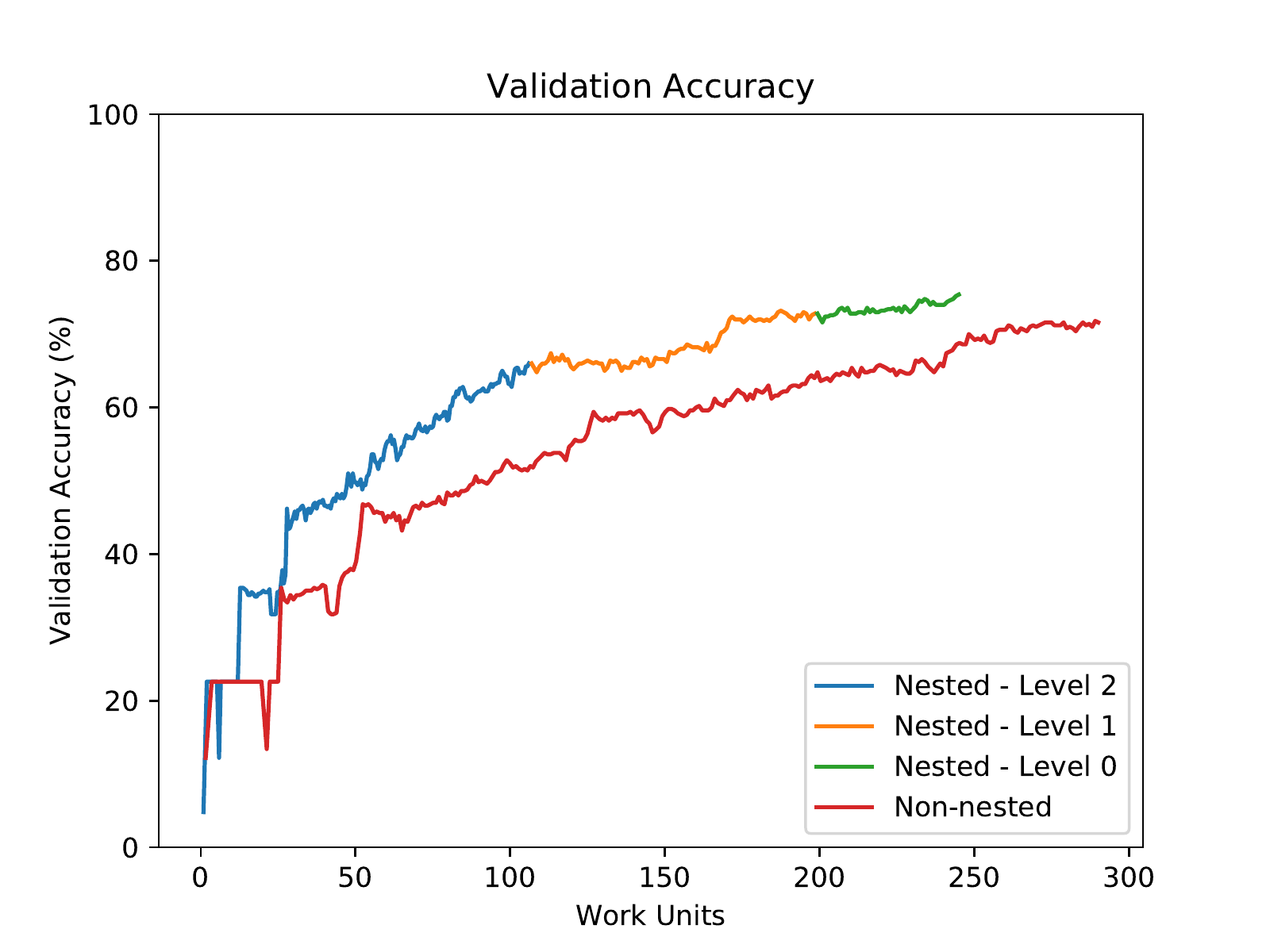}}
\centering{
\includegraphics[width=0.4\textwidth]{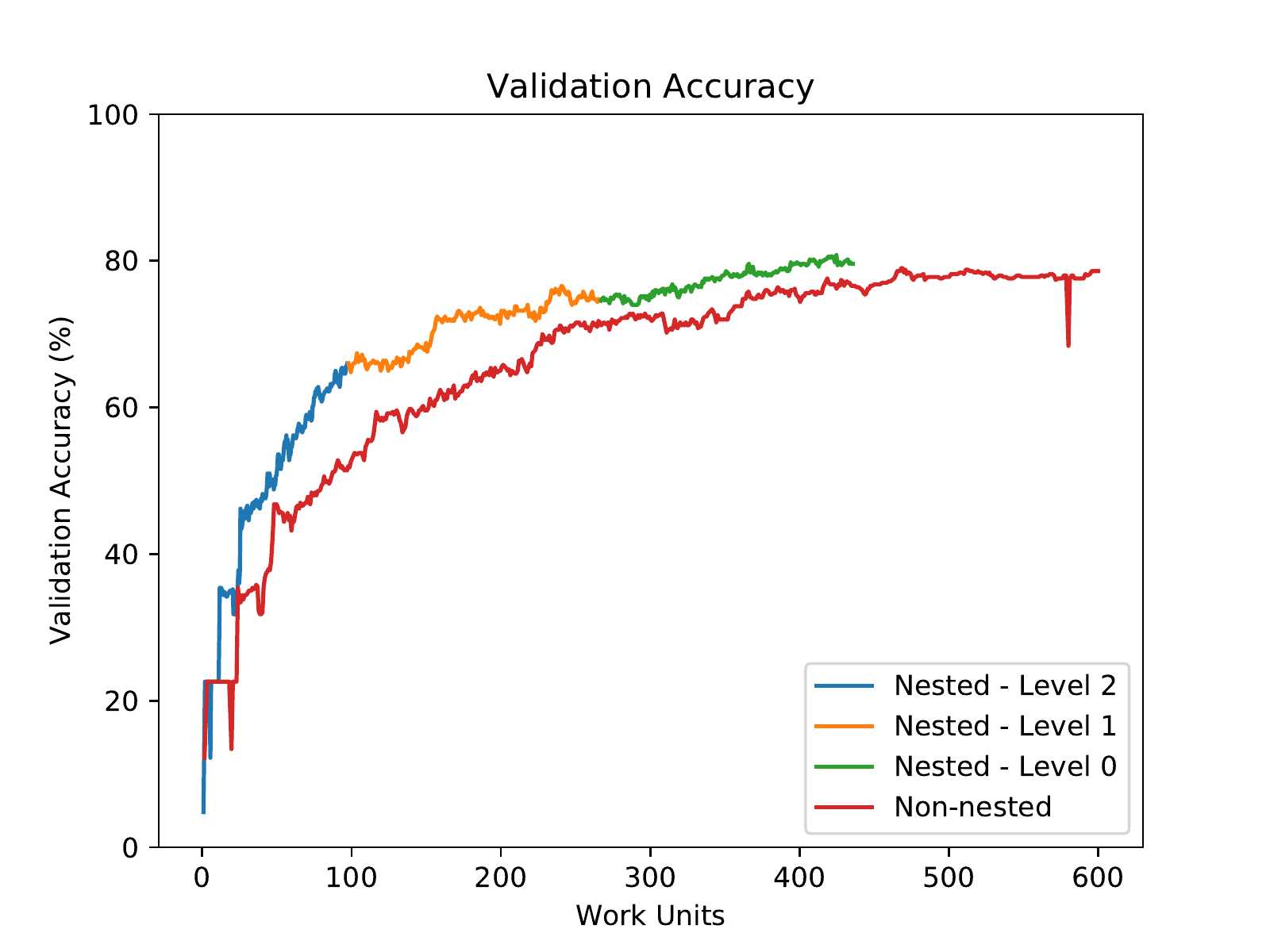}}
\caption{Validation accuracy as a function of computational work units for two representative algorithmic configurations of the nested iteration algorithms. This compares the relative run time of the non-nested iteration to the run time of the nested-iteration.
In the top image a schedule with $200$, $100$, and $50$ optimization iterations on the coarse to the fine level is used by nested iteration. The bottom image has uses $200$ iterations regardless of level. \label{fig:ip-train-images}}
\end{figure}

Figure~\ref{fig:ip-train-images} compares the validation accuracy of training with the non-nested algorithm to two nested iteration strategies. For all cases, the validation accuracy is plotted as a function of work units performed throughout the optimization solver. The first nested iteration strategy uses a schedule that performs $m^{(2)}=200$ optimization iterations on the coarse level, $m^{(1)}=100$ on the medium, and just $m^{(0)}=50$ iterations on the fine. This approach is designed to reduce the run time assuming that most of the work of training can be done on coarser levels. Note, as a result of not iterating on the fine grid as much, this may result in a lower achieved validation accuracy (indeed this is born out by the results). The top plot in the figure shows the non-nested iteration (red line), and the three different levels of the nested iteration (multiple colors). Considering only the coarse problem, it is clear that the nested iteration achieves higher validation accuracy in less computational time. This can be attributed to the much greater number of iterations taken. Where the increase in speed is a result of a shallower network. Moreover considering the entire run, the nested iteration has larger validation accuracy after just $25$ work units.

The lower image shows a similar story. However, this time the schedule for the nested iteration uses $m^{(2)}=m^{(1)}=m^{(0)}=200$ optimization steps at each level. Here again its clear that training for many steps on the coarse level yields rapid improvements. Overall, higher validation accuracy is still achieved in all cases for a fixed number of work units. Relative to the previous schedule (top image), the validation accuracy of this uniform schedule is improved, though at the cost of longer training times. 

\section{Conclusion}
In this work, a nested iteration strategy for network initialization that enhances recent advances in layer-parallel training methodologies is developed for ODE networks. This approach uses a training algorithm where a sequence of neural networks are successively trained for a geometrically increasing number of layers. Interpolation operators are defined that transfer the weights between the levels. Results presented for the Peaks and Indian Pines classification example problems show that nested iteration can achieve greater accuracy at less computational cost. An exciting additional benefit was observed for the peaks problem. In this case the nested iteration also provided a structural regularization effect that resulted in reduced variation over repeated runs in a hyperparameter sweep. A more thorough investigation of this result, and greater improvements to the nested iteration and layer-parallel algorithms are the subject of future work.

\section*{Acknowledgements}
The work of E. C. Cyr was supported by Sandia National Laboratories and the DOE Early Career Research Program.
Sandia National Laboratories is a multimission laboratory managed and operated by National Technology \& Engineering Solutions of Sandia, LLC, a wholly owned subsidiary of Honeywell International Inc., for the U.S. Department of Energy's National Nuclear Security Administration under contract DE-NA0003525. The views expressed in the article do not necessarily represent the views of the U.S. Department of Energy or the United States Government.
S. G\"{u}nther was supported by Lawrence Livermore National Laboratory.
This work was performed under the auspices of the U.S. Department of Energy by Lawrence Livermore National Laboratory under contract DE-AC52-07-NA27344. LLNL-PROC-798920.

This document was prepared as an account of work sponsored by an agency of the United States government. Neither the United States government nor Lawrence Livermore National Security, LLC, nor any of their employees makes any warranty, expressed or implied, or assumes any legal liability or responsibility for the accuracy, completeness, or usefulness of any information, apparatus, product, or process disclosed, or represents that its use would not infringe privately owned rights. Reference herein to any specific commercial product, process, or service by trade name, trademark, manufacturer, or otherwise does not necessarily constitute or imply its endorsement, recommendation, or favoring by the United States government or Lawrence Livermore National Security, LLC. The views and opinions of authors expressed herein do not necessarily state or reflect those of the United States government or Lawrence Livermore National Security, LLC, and shall not be used for advertising or product endorsement purposes. 




\bibliographystyle{spmpsci}      
\bibliography{refs}   

\end{document}